\documentclass{article}

\usepackage{nac_preprint}

\usepackage[utf8]{inputenc} 
\usepackage[T1]{fontenc}    
\usepackage{hyperref}       
\usepackage{url}            
\usepackage{booktabs}       
\usepackage{amsfonts}       
\usepackage{nicefrac}       
\usepackage{microtype}      

\usepackage{booktabs} 
\usepackage{subfig}
\usepackage{amsmath}
\usepackage{mathtools}
\usepackage[toc,page]{appendix}
\usepackage{enumitem}
\usepackage{graphics}
\usepackage{graphicx}
\usepackage{algorithm}
\usepackage[noend]{algorithmic}
\usepackage{color}
\usepackage{dsfont}
\DeclareMathOperator*{\argmin}{arg\,min}

\newcommand{\etal}{{\it et al.}}

\title{Reducing Catastrophic Forgetting in Self Organizing Maps with Internally-Induced Generative Replay}

\author{%
  Hitesh Vaidya, Travis Desell, Alexander Ororbia \\
  Rochester Institute of Technology \\
  \{\texttt{hv8322}, \texttt{tjdvse}\}\texttt{@rit.edu},
    \texttt{ago@cs.rit.edu}
}

\begin{document}
\setlength{\abovedisplayskip}{0.065cm}
\setlength{\belowdisplayskip}{0pt}

\maketitle

\begin{abstract}
\label{sec:abstract} 
A lifelong learning agent is able to continually learn from potentially infinite streams of pattern sensory data. One major historic difficulty in building agents that adapt in this way is that neural systems struggle to retain previously-acquired knowledge when learning from new samples. This problem is known as catastrophic forgetting (interference) and remains an unsolved problem in the domain of machine learning to this day. 
While forgetting in the context of feedforward networks has been examined extensively over the decades, far less has been done in the context of alternative architectures such as the venerable self-organizing map (SOM), an unsupervised neural model that is often used in tasks such as clustering and dimensionality reduction. Although the competition among its internal neurons might carry the potential to improve memory retention, we observe that a fixed-sized SOM trained on task incremental data, i.e., it receives data points related to specific classes at certain temporal increments, experiences significant forgetting.
In this study, we propose the continual SOM (c-SOM), a model that is capable of reducing its own forgetting when processing information. 
\end{abstract}

\section{Introduction}
\label{sec:intro}

Lifelong machine learning, otherwise known as continual and never-ending learning \cite{chen2016lifelong}, stands as one of the greatest challenges facing artificial intelligence research 
, especially with respect to models parameterized by deep neural networks (DNNs). In this problem context, an agent must attempt to learn not just one single prediction task using one single dataset, but rather, it must learn online \emph{across} several task datasets, much as human agents do, aggregating and transferring its knowledge as new pattern vectors are encountered.
DNNs particularly struggle to learn continually due to the well-known classical fact that they tend to \emph{catastrophically forget} \cite{french1999catastrophic}, or rather, they completely erase the knowledge acquired during the learning of earlier tasks when processing samples from new tasks.

While catastrophic forgetting has been explored extensively in DNNs, very little work has focused on the occurrence and reduction of forgetting in less popular architectures, such as self-organizing maps (SOMs) \cite{kohonen1982self}, especially in the context of unsupervised problem scenarios.
SOMs are a type of competitive neural system where internal neurons compete for the right to activate in the presence of a data pattern and synaptic parameters 
are adjusted through Hebbian learning \cite{hebb1949organization,martinetz1993competitive}. Neuronal units can be arranged in a topological fashion, i.e., a neighbourhood based on Euclidean distance between the activation values of units, as well as a spatial fashion, i.e., units arranged in Cartesian plane. Given that units in a competitive learning system like the SOM tend to specialize for certain types of patterns (serving as data prototypes), such a system would appear to be suited for combating catastrophic forgetting.
This would especially appear to be the case, since, in theory, competing units would lead to reduced neural cross-talk \cite{ororbia2021continual}, which has been shown to be a key source of forgetting \cite{mccloskey_catastrophic_1989,ratcliff_connectionist_1990,french1999catastrophic}. 
Furthermore, approaches, such as that presented by Gepperth \etal~ \cite{7346155}, have advocated the use of SOMs for continual learning. However, the SOM in its purest form, as we observe in our experiments, is itself prone to forgetting \cite{frontiersForgetting}. 

\begin{figure}[t]
    \centering
    \includegraphics[width=0.675\textwidth]{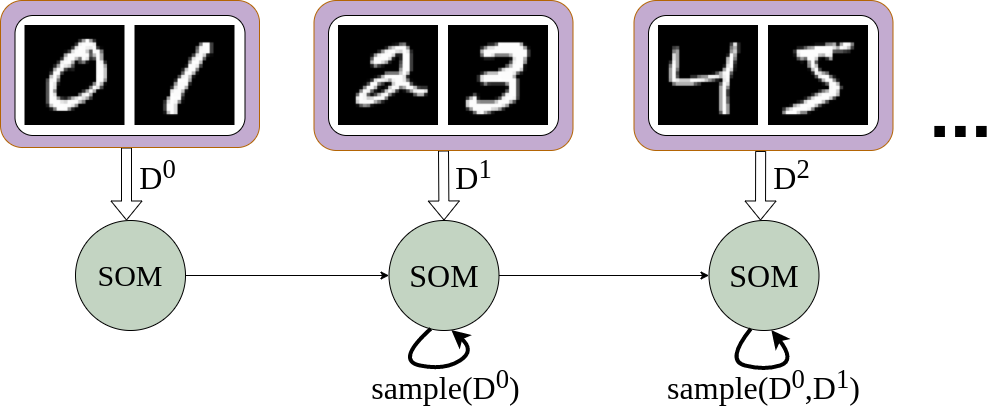}
    \caption{The proposed continual SOM (c-SOM) neural system processing a sequence of tasks, each containing two different classes of digits. $\mathcal{D}^t$ denotes the $t$th dataset associated with task $t$ and $\text{sample}(\mathcal{D}^{t-k})$ (where $k = 1, 2,...$) indicates that the c-SOM samples its units to synthesize patterns for any task $t-k$ in order to refresh its memory and mitigate forgetting while learning from the current task $t$.}
    \label{fig:SOM_replay}
\end{figure}

In this work (Figure \ref{fig:SOM_replay}), we propose the continual SOM (c-SOM), an SOM neural model that actively reduces the amount of forgetting that it experiences when incrementally learning from data. Specifically, we modify the SOM's decay function to be task-dependent and extend its units to self-induce a form of generative rehearsal/replay to improve memory retention.


\section{Methodology}
\label{sec:methods}

\paragraph{Problem Setup:} Consider a sequence of $N$ tasks, denoted by $\mathcal{S} = \bigcup_{t=1}^T \mathcal{T}_t$. Each task $\mathcal{T}_t$ has a training dataset (with $C$ classes), $\mathcal{D}_{train}^{(t)} = \bigcup_{i=1}^{N_t} \{(\mathbf{x}_i^{(t)}, \mathbf{y}_i^{(t)})\}$, where $\mathbf{x}_i^{(t)} \in \mathcal{R}^{D \times 1}$ is a data pattern and  $\mathbf{y}_i^{(t)} \in \{0,1\}^{C \times 1}$ is its label vector ($N_t$ is the number of data points in task $T_t$). 
Similarly, $\mathcal{D}_{test}^{(t)}$ is the test dataset for task $\mathcal{T}_t$. 
When a lifelong learning model is finished training on task $\mathcal{T}_t$ using $\mathcal{D}_{train}^{(t)}$, $\mathcal{D}_{train}^{(t)}$ will be lost as soon as the model proceeds to task $\mathcal{T}_{t+1}$ to $\mathcal{T}_T$ ($\mathcal{D}_{test}^{(t)}$ is only used for evaluation). 
The objective is to maximize the agent's performance 
on task $\mathcal{T}_t$ while minimizing how much its performance on prior tasks $\mathcal{T}_1$ to $\mathcal{T}_{t-1}$ degrades. 

\paragraph{The Model:} For our c-SOM (Algorithm \ref{alg:som_training}), $\sigma$, $\lambda$ are, respectively, the initial radius and learning rate values. $\sigma_t$, $\lambda_t$ are their values for task $\mathcal{T}_t$. Both values decay as follows: 
\begin{align}
    \label{eqn:decay_lr}
    \lambda_t &= \lambda (1 + t * \exp{(t / \tau_\lambda)})^{-1} \\
    \sigma_t &= \sigma (1 + t * \exp{(t / \tau_\sigma)})^{-1} \mbox{.}
\end{align}
$\tau_\lambda = \tau_\sigma = 1000$ are the time constants. 
$u$ is the best matching unit (BMU) while $v$ is any other unit in the SOM (with $m$ units). 
$h(u, v, t)$ represents the neighbourhood function between units $u$ and $v$ for task $\mathcal{T}_t$. 
Each unit $v$ in our SOM is composed of two coupled vectors -- its prototype weights $\mathbf{w}_v \in \mathcal{R}^{D \times 1}$
and its running variance weights $\mathbf{w}^\sigma_v \in \mathcal{R}^{D \times 1}$, where $\mathbf{w}_v$ is updated via a Hebbian update \cite{martinetz1993competitive} and $\mathbf{w}^\mu_v$ and $\mathbf{w}^\sigma_v$ are updated  via Welford's algorithm for calculating variance \cite{welford1962note}.
$\phi(c)$ returns the number of units trained on class $c$ (out of $C$)
, while $\rho_v$ stores the class that unit $v$ maps to. At each simulation step, the c-SOM generates $K$ samples from $m_r$ randomly chosen trained units via: $\mathbf{s}^v = \mathbf{w}_v + \sqrt{\mathbf{w}^\sigma_v} \odot \mathbf{\epsilon}$ ($\mathbf{\epsilon} \sim \mathcal{N}(0,1)$). The $K \times m_r$ samples are then used to refresh the c-SOM (via the same update rules).

\begin{algorithm}[!t]
\caption{The c-SOM Training Process}
\label{alg:som_training}
\textbf{Input}: Task input, $\mathbf{x}_i^{(t)}$, c-SOM weight parameters $\{\mathbf{w}_v, \mathbf{w}^\sigma_v \}$ \\
\textbf{Parameter}: $\sigma, \lambda$
\begin{algorithmic}[1] 
\FOR{$t=0$ to $T$}
    \STATE{$\sigma_t, \lambda_t \gets$ decay parameters for t} // (Eqn 1 \& 2)
    \STATE Update $\mathbf{w}_v$ on ($\mathbf{x}_i^{(t)}$, $\sigma_t$, $\lambda_t$) // (Hebbian update for the prototypes)
    \STATE Update  $\mathbf{w}^\sigma_v$ (for all units) // (Welford update for the variance parameters)
    \STATE Generate $K$ samples from $m_r$ randomly chosen (trained) units \& retrain c-SOM on these
\ENDFOR
\end{algorithmic}
\end{algorithm}

\begin{figure}[!ht]
    \centering
    \includegraphics[width=0.385\textwidth]{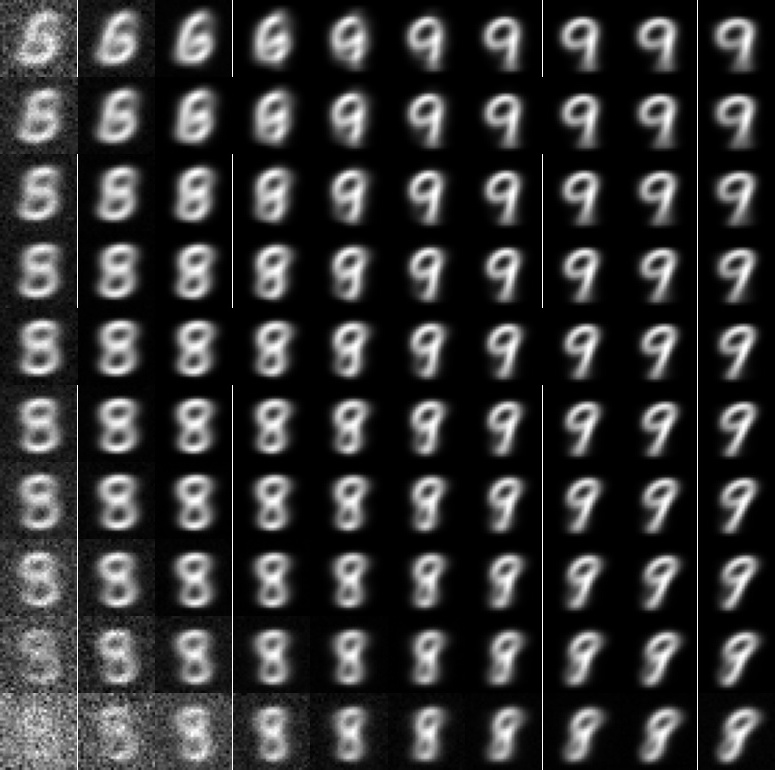}
    \hspace{0.35cm}
    \includegraphics[width=0.385\textwidth]{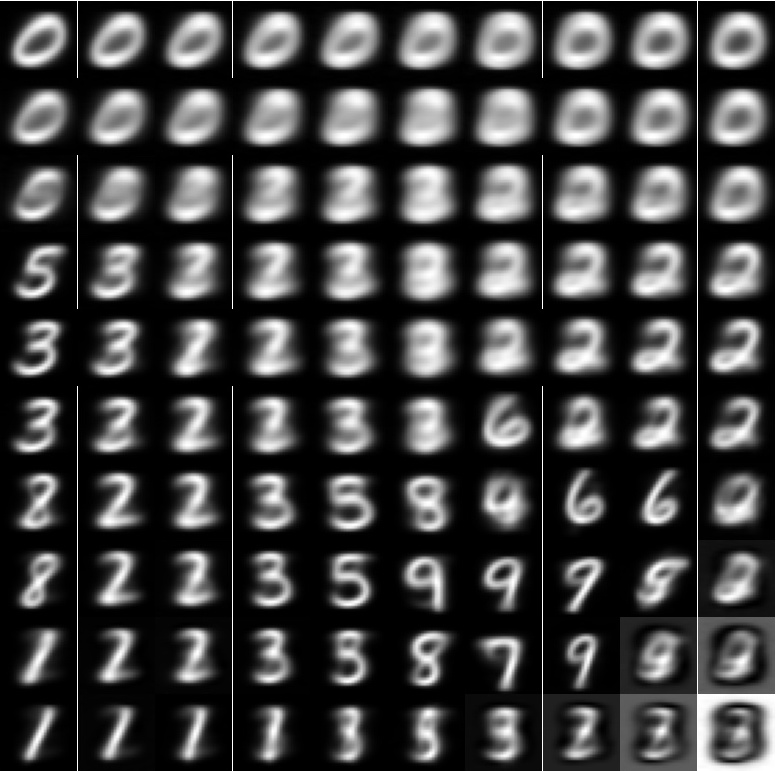}
    \caption{Classical SOM (Left) versus the proposed c-SOM (Right). Both models were trained on Split-MNIST in a class incremental fashion (yielding $10$ tasks in total, one per uniqe digit).}
    \label{fig:som_comparison}
\end{figure}

Figure \ref{fig:som_comparison} (Left) displays the final output after training a classical SOM on the benchmark Split-MNIST
($C = 1$ per task)
with a grid shape of $10 \times 10$ (this small grid size was chosen to exacerbate forgetting). 
Note that Split-MNIST contains $28\times28$ images with gray-scale pixel values, i.e., range is $[0,255]$, each coming from one of ten possible classes (digits $0$ through $9$).
As seen among its prototypes, this SOM remembered only classes $6$, $8$ and $9$. Except for $6$, classes $8$ and $9$ are the ones encountered in the last task of Split-MNIST. 
In contrast, Figure \ref{fig:som_comparison} (Right) shows our proposed c-SOM with the same shape. Desirably, our model contains units that represent digits in all classes encountered across all tasks of Split-MNIST, exhibiting far less forgetting. Note that all SOMs were trained in a \emph{class incremental} fashion.

\begin{algorithm}[!t]
\caption{The $\alpha_{\text{mem}}$ Metric}
\label{alg:alpha_metric}
\textbf{Input}: $\mathbf{X}_{test}$, where $\mathbf{X}^{c}_{test}$ is all $\mathbf{x}_i$ w/ $\arg\max(\mathbf{y}_i) \equiv c$\\
\textbf{Parameter}: Weights $\{\mathbf{w}_v, \mathbf{w}^\sigma_v\}$ 
\begin{algorithmic}[1] 
\STATE{$N_c = V / C$, where C = total number of classes}
\STATE{$\rho_v := \argmin \big( d = \{\|\mathbf{w}_v - \mathbf{X}_{test}^{(c)}\| : \forall \ c \ \epsilon \ C\}$} \big)
\STATE{$\phi = \left\{\sum_{v=0}^V \mathds{1}(c = \rho_v) : \forall \ c \ \epsilon \ C \right\}$}
\STATE{$\alpha_{\text{mem}} = (\phi_i - N_c)_{RMS}$} // RMS = ``root-mean-square''
\end{algorithmic}
\end{algorithm}

In order to measure the performance of the incrementally learned SOMs, we designed a novel metric, $\alpha_{\text{mem}}$ (see Algorithm \ref{alg:alpha_metric}). 
An ideal lifelong learning SOM, assuming equal variance among classes and tasks,
would have $\alpha_{\text{mem}}=0$, which would indicate an equal representation of classes across units. Therefore, lower $\alpha_{\text{mem}}$ values represent better SOM performance. We experimented with different parameter settings of ($\sigma, \lambda$) for three different SOM versions -- a classical SOM and two c-SOM variants. The $\sigma$ value was selected out of \{2,3,4,5\} while the $\lambda$ was chosen from \{0.001, 0.001, 0.007, 0.01\}. 
Table \ref{tab:results} presents the best $\alpha_{\text{mem}}$ value along with their corresponding mean and standard deviations over $10$ trials.

\begin{table}[!t]
    \centering
    \begin{tabular}{c|c|c|c|c}
        Model & $(\sigma, \lambda)$ & $\alpha_{mem}$ & $\mu_{\alpha_{mem}}$ & $\sigma_{\alpha_{mem}}$ \\
        \hline
        SOM & (2,0.001) & 12.75 & 17.97 & 3.02  \\ 
        c-SOM (K=1) & (3,0.01) & 11.85 & 12.49 & 0.45 \\ 
        \textbf{c-SOM (K=2)} & \textbf{(3,0.01)} & \textbf{9.9} & \textbf{12.15} & \textbf{1.25} 
    \end{tabular}
    \caption{Parameters with best (minimum) $\alpha_{\text{mem}}$ values for each model ($m_r = 1$). The mean and standard deviation for $\alpha_{mem}$, i.e., $\mu_{\alpha_{mem}}$ and  $\sigma_{\alpha_{mem}}$, over $10$ trials are reported for the Split-MNIST benchmark.}
    \label{tab:results}
\end{table}

\section{Conclusion}
\label{sec:conclusion}
In this study, we proposed a variant of the SOM that actively reduces the amount of forgetting it experiences when trained in a class-incremental fashion. Qualitative and quantitative results on Split-MNIST demonstrate a desirable improvement in memory retention.

\bibliographystyle{acm}
\bibliography{ref}

\end{document}